\documentclass[conference]{IEEEtran}
\IEEEoverridecommandlockouts
\usepackage{cite}
\usepackage{amsmath,amssymb,amsfonts}
\usepackage{algorithm}
\usepackage{algorithmic}
\usepackage{graphicx}
\usepackage{textcomp}
\usepackage{xcolor}
\usepackage{booktabs}
\usepackage{url}
\def\BibTeX{{\rm B\kern-.05em{\sc i\kern-.025em b}\kern-.08em
    T\kern-.1667em\lower.7ex\hbox{E}\kern-.125emX}}
\begin{document}

\title{SURE-RAG: Sufficiency and Uncertainty-Aware Evidence Verification for \\Selective Retrieval-Augmented Generation}

\author{\IEEEauthorblockN{Jingxi Qiu\textsuperscript{1,2}, Zeyu Han\textsuperscript{2,3}, Cheng Huang\textsuperscript{1,3,$\dag$}}
\IEEEauthorblockA{\textit{\textsuperscript{1}ZenWeave AI, \textsuperscript{2}Georgetown University} \\
\textit{\textsuperscript{3}Southern Methodist University}\\
\{jingxi, chenghuang\}@zenweaveai.com \\
$\dag$ Corresponding Author}
}

\maketitle

\begin{abstract}
Retrieval-augmented generation (RAG) grounds answers in retrieved passages, yet relevance does not guarantee sufficiency: a topical passage may still fail to justify the answer. We study evidence sufficiency verification for selective RAG answering, in which a verifier receives a question, a candidate answer, and retrieved evidence and decides whether the evidence supports, refutes, or is insufficient for the answer, answering only when support is established.

We present SURE-RAG, an aggregation protocol that treats evidence sufficiency as a set-level property: missing hops and unresolved conflicts cannot be detected by scoring passages independently. A shared claim-evidence verifier produces a local relation distribution for each (claim, passage) pair, which SURE-RAG aggregates into four interpretable answer-level feature blocks (coverage, relation strength, uncertainty, and retrieval), producing a three-way decision and an auditable selective score. We evaluate on HotpotQA-RAG v3, a controlled multi-hop benchmark, under an artifact-aware protocol (shortcut baselines, counterfactual swaps, no-oracle checks, and GPT-4o audits). Calibrated SURE-RAG attains 0.9075 Macro-F1 (raw 0.8951 $\pm$ 0.0069), well above DeBERTa mean-pooling (0.6516) and a GPT-4o judge (0.7284), and on par with a strong concat cross-encoder (0.8888 $\pm$ 0.0109) while remaining fully auditable. At 30\% coverage, risk falls from 0.2588 to 0.1642, a 37\% relative reduction. As a boundary-mapping experiment, we contrast SURE-RAG with GPT-4o on HaluBench unsafe detection: the ranking reverses (0.3343 vs.\ 0.7389 unsafe-F1), indicating that controlled sufficiency verification and natural hallucination detection are distinct problems.
\end{abstract}


\section{Introduction}

Retrieval-augmented generation (RAG) improves factual grounding by conditioning language model outputs on retrieved passages \cite{lewis2020rag}. Relevance, however, is not sufficiency: a passage can mention the right entities or intermediate facts yet fail to justify the answer, and the retrieved set can be incomplete or contradictory. A reliable RAG system therefore needs a decision layer that asks not whether evidence is relevant, but whether it suffices to justify the answer.

\begin{figure}[!t]
\centering
\includegraphics[width=0.85\columnwidth]{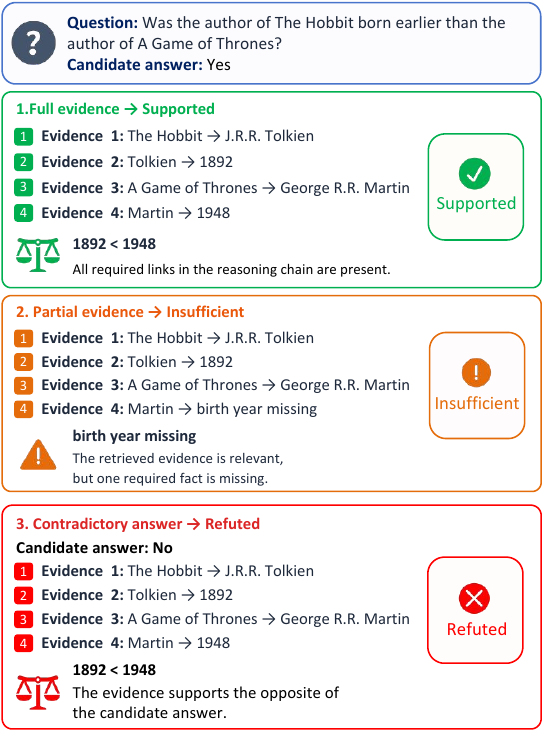}
\caption{Three multi-hop evidence conditions for the same question and candidate answer. With full evidence the answer is \textsc{Supported}; removing one required hop yields \textsc{Insufficient}; perturbing the candidate answer against the original supporting evidence yields \textsc{Refuted}. Evidence sufficiency is therefore a set-level property of the retrieved context, not a function of any single passage.}
\label{fig:evidence-conditions}
\end{figure}

We study \emph{evidence sufficiency verification} for selective RAG answering. Given a question $q$, an answer $a$, and retrieved evidence $E$, the verifier assigns one of three labels (\textsc{Supported}, \textsc{Refuted}, or \textsc{Insufficient}), of which only \textsc{Supported} permits answering. The three labels matter because missing, partial, and contradicting evidence differ in meaning even though they share the same answer-or-abstain consequence, a distinction that binary hallucination detection collapses. Figure~\ref{fig:evidence-conditions} illustrates the three conditions on a shared multi-hop question.

A standard verifier scores each passage independently and pools the scores with a max or mean operator. Both are brittle for sufficiency: max-pooling over-answers when one passage looks supportive but a required hop is missing, while mean-pooling dilutes a decisive refutation among neutral passages. Sufficiency is a set-level property of the retrieved context, not a local entailment score.

We propose \textbf{SURE-RAG}, an aggregation protocol that maps pair-level relation distributions to answer-level sufficiency decisions. A shared verifier scores each (claim, passage) pair as support, refute, or neutral; SURE-RAG aggregates these scores into four interpretable feature blocks (coverage, relation strength, uncertainty, and retrieval) and emits a calibrated decision and a selective score. Because SURE-RAG and the pooling baselines share this verifier, our comparison isolates the effect of aggregation.

Synthetically constructed sufficiency benchmarks can encode shortcuts (answer-style leakage, length artifacts, lexical overlap, or construction-derived retrieval metadata) that inflate apparent performance. Our evaluation therefore pairs each model with shortcut baselines, counterfactual evidence swaps, no-oracle retrieval checks, GPT-4o semantic audits and judge comparisons, and risk-coverage metrics.

On HotpotQA-RAG v3, a controlled multi-hop benchmark with five evidence conditions (full, partial, hard-insufficient, irrelevant, and naturally refuted), calibrated SURE-RAG reaches 0.9075 Macro-F1 across three neural seeds, with raw SURE-RAG at 0.8951 $\pm$ 0.0069. This is far above DeBERTa mean-pooling (0.6516) and on par with a strong concat cross-encoder (0.8888 $\pm$ 0.0109), while retaining block-level auditability. Selective answering further lowers unsafe-answer risk at low coverage. To map the task boundary, we contrast SURE-RAG with GPT-4o on HaluBench unsafe detection; here the ranking reverses, showing that the two tasks reward different inductive biases.

Our contributions are:
\begin{itemize}
    \item We formulate evidence sufficiency verification as a selective RAG answering problem with \textsc{Supported}, \textsc{Refuted}, and \textsc{Insufficient} labels.
    \item We propose SURE-RAG, an auditable answer-level aggregation protocol over local claim-evidence relation and uncertainty features.
    \item We introduce an artifact-aware evaluation protocol with shortcut baselines, counterfactual swaps, no-oracle checks, GPT-4o audits, and risk-coverage metrics.
    \item On HotpotQA-RAG v3, SURE-RAG clearly outperforms pair-level pooling baselines, matches a strong concat cross-encoder while remaining auditable, and improves low-coverage selective-answering risk.
    \item We use HaluBench unsafe detection as a deliberate boundary-mapping experiment that separates controlled sufficiency verification from natural hallucination detection.
\end{itemize}

\section{Related Work}

\subsection{RAG, Fact Verification, and Evidence Sufficiency}

RAG conditions language model outputs on retrieved evidence \cite{lewis2020rag}, with dense retrieval \cite{karpukhin2020dpr} and reader-generator architectures \cite{izacard2021fid} further improving retrieval-conditioned QA. These systems treat retrieval as context for generation; we instead treat retrieved passages as evidence subject to sufficiency verification before answering.

Fact verification benchmarks provide the closest label structure. FEVER labels claims as supported, refuted, or not enough information against Wikipedia evidence \cite{thorne2018fever}; SciFact adapts the same distinction to scientific claims with rationale annotations \cite{wadden2020scifact}. Later benchmarks broaden the evidence setting to mixed structured and unstructured sources \cite{aly2021feverous} and to contrastive, revision-based evidence designed to resist lexical shortcuts \cite{schuster2021vitaminc}. SURE-RAG inherits this three-way distinction but operates in a different decision setting: a candidate RAG answer is verified against retrieved passages and routed to answer or abstain. Our SciFact pilot further shows that naively reusing such benchmarks for sufficiency variants can be shortcut-prone, motivating the artifact-aware protocol described below.

\subsection{RAG Evaluation and Hallucination Detection}

Two complementary lines evaluate RAG trustworthiness: RAGAS and ARES measure system-level dimensions such as context relevance, answer faithfulness, and answer relevance \cite{es2024ragas,saadfalcon2024ares}, while SelfCheckGPT, RAGTruth, and HaluBench address context-grounded hallucination through self-consistency probing or labeled data \cite{manakul2023selfcheckgpt,niu2024ragtruth,patronusai2024halubench}. SURE-RAG addresses an orthogonal question: whether retrieved evidence \emph{suffices} to justify the answer, a three-way distinction that binary hallucination labels cannot express. Our HaluBench experiments confirm this orthogonality directly: SURE-RAG and a GPT-4o judge reverse rank across the two task settings.

\subsection{Attribution, Long-Form Factuality, and Selective Prediction}

Citation and attribution work asks whether generated claims are supported by cited sources. ALCE evaluates citation-aware generation, and FActScore decomposes long-form generations into atomic facts to score source-supported factuality \cite{gao2023alce,min2023factscore}. SURE-RAG adopts the same claim-level view, but our main experiments target short-answer multi-hop sufficiency rather than long-form claim coverage.

Rationale evaluation surfaces a parallel methodological concern. RORA shows that free-text rationales can paraphrase the target label rather than provide substantive justification, inflating apparent explanation quality \cite{jiang2024rora}. Although RORA targets generated rationales while we evaluate retrieved passages, both lines reject apparent support driven by label leakage, lexical overlap, or construction artifacts rather than evidential sufficiency.

Selective prediction studies when a model should abstain instead of predicting \cite{chow1970reject,geifman2017selective}, with calibration and conformal prediction providing tools for confidence estimation and risk control \cite{guo2017calibration,vovk2005algorithmic,angelopoulos2021conformal}. RAG introduces evidence-specific uncertainty sources (missing evidence, retrieval uncertainty, evidence disagreement, and support-refute conflict), which SURE-RAG aggregates into a selective score evaluated under risk-coverage metrics.
\section{Problem Formulation}

A RAG system receives a question $q$, retrieves evidence passages $E=\{e_1,\ldots,e_k\}$, and produces an answer $a$. We ask whether $E$ suffices to justify $a$.

\subsection{Evidence Sufficiency Verification}

The verification task predicts $y\in\{\textsc{Supported},\textsc{Refuted},\textsc{Insufficient}\}$ from $(q,a,E)$. \textsc{Supported} means $E$ entails the central factual content of $a$; \textsc{Refuted} means $E$ contradicts at least one central factual claim of $a$; \textsc{Insufficient} covers all remaining cases in which $E$ cannot justify $a$: missing hops, partial evidence, topical-but-relation-missing passages, or unresolved internal conflict.

This distinction matters because relevance does not imply sufficiency. A passage may mention the correct entity without establishing the required relation; a multi-hop question may retrieve one supporting hop while missing another. Either case can make answering unsafe despite superficially successful retrieval.

\subsection{Selective Answering}

For selective answering, \textsc{Supported} maps to \textsc{Safe} and the other two labels map to \textsc{Unsafe}. The system outputs $d\in\{\textsc{Answer},\textsc{Abstain}\}$; an answer is \emph{unsafe} when the underlying example is not \textsc{Supported}. Given a selective score $s$, threshold $\tau$, and answered set $A_\tau=\{i:s(q_i,a_i,E_i)\geq\tau\}$, we define
\begin{align}
\mathrm{Coverage}(\tau)&=|A_\tau|/N,\\
\mathrm{Risk}(\tau)&=\frac{\#\,\text{unsafe answers in }A_\tau}{|A_\tau|}.
\end{align}
We report risk at fixed coverage and, when applicable, coverage at fixed risk as a diagnostic, with thresholds selected on the development split.

\subsection{Claim-Level Extension}

SURE-RAG is formulated at the claim level: an answer is decomposed into claims $C(a)=\{c_1,\ldots,c_m\}$, with $m=1$ for typical short answers and $m>1$ for long-form generations. Long-form sufficiency then depends on per-claim verification, claim coverage, and inter-claim contradiction. Our main experiments validate short-answer multi-hop sufficiency; full long-form claim coverage is left to future work.

\begin{figure*}[!t]
\centering
\includegraphics[width=0.9\textwidth]{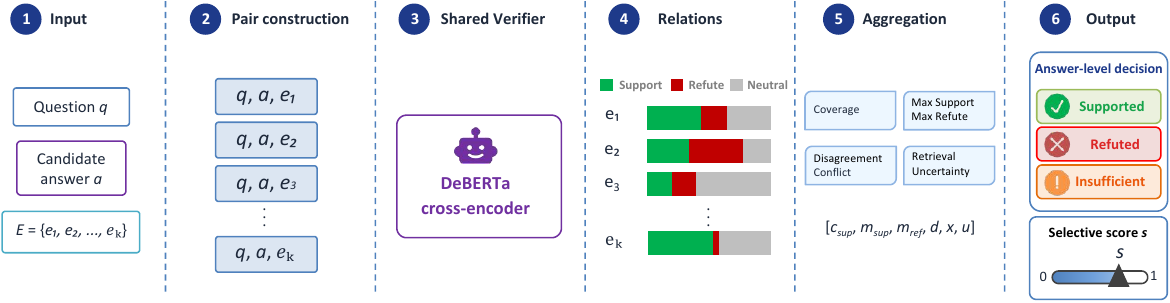}
\caption{The SURE-RAG pipeline. Given the input triple $(q, a, E)$ with $|E|=k$ retrieved passages (step~1), each passage is paired with the question and candidate answer (step~2) and scored by a shared DeBERTa cross-encoder (step~3) into a local relation distribution over support, refute, and neutral, visualized as stacked bars (step~4). The pair-level relation matrix is then aggregated (step~5) into answer-level features: coverage, max-support, max-refutation, disagreement, conflict, and retrieval uncertainty. A lightweight classifier $g_\phi$ produces the three-way decision $\hat{y} \in \{\textsc{Supported}, \textsc{Refuted}, \textsc{Insufficient}\}$ together with the selective score $s = \pi_{\textsc{Supported}} - \beta u$ (step~6), in which $u$ is the aggregate uncertainty penalty; the system answers when $\hat{y} = \textsc{Supported}$ and $s \geq \tau$, and otherwise abstains.}
\label{fig:sure-rag-pipeline}
\end{figure*}

\section{Method}
SURE-RAG comprises three components: pair-level claim-evidence verification, answer-level sufficiency aggregation, and selective answering, summarized in Figure~\ref{fig:sure-rag-pipeline} and Algorithm~\ref{alg:sure-rag}. The first scores each (claim, passage) pair locally; the second turns these local scores into an answer-level feature vector; the third combines the resulting label distribution with an uncertainty penalty to decide whether to answer.

\subsection{Claim-Evidence Verification}

For each claim $c_j \in C(a)$ and each evidence passage $e_i \in E$, a DeBERTa-v3-base cross-encoder \cite{he2023debertav3} predicts a local relation distribution
\begin{equation}
\mathbf{p}_{ij}=h_\theta(q,c_j,e_i)=[p_{ij}^{\mathrm{sup}},p_{ij}^{\mathrm{ref}},p_{ij}^{\mathrm{neu}}],
\label{eq:pair-verifier}
\end{equation}
giving supported, refuted, and neutral probabilities. The same pair verifier is shared with all neural pooling baselines, so any improvement of SURE-RAG over those baselines reflects answer-level aggregation rather than a stronger encoder. For long-form answers, the claim set $C(a)$ can be produced by FActScore-style atomic-claim decomposition \cite{min2023factscore}; our main experiments use single-claim short answers.

\subsection{Answer-Level Sufficiency Aggregation}

Where pooling baselines either over-commit (max) or dilute (mean), SURE-RAG summarizes the full set of pair-level distributions $\{\mathbf{p}_{ij}\}$ into interpretable feature blocks (Table~\ref{tab:feature-blocks}) and predicts
\begin{equation}
P(y\mid q,a,E)=g_\phi(\mathbf{z}(q,a,E)),
\label{eq:answer-classifier}
\end{equation}
where $y\in\{\textsc{Supported},\textsc{Refuted},\textsc{Insufficient}\}$ and $\mathbf{z}$ is the answer-level feature vector. The aggregator $g_\phi$ is a logistic-regression classifier (readily replaceable with other models) trained on the training split.

\begin{table}[t]
\flushleft
\small
\setlength{\tabcolsep}{3pt}
\caption{Answer-level feature blocks aggregated by SURE-RAG from pair-level distributions $\{\mathbf{p}_{ij}\}$. The retrieval block is optional and is removed or replaced in no-oracle checks.}
\label{tab:feature-blocks}
\begin{tabular}{p{0.18\columnwidth}p{0.30\columnwidth}p{0.18\columnwidth}p{0.20\columnwidth}}
\toprule
Block & Features & Source over $\{\mathbf{p}_{ij}\}$ & Purpose \\
\midrule
Coverage & supported, refuted, insufficient claim fractions & per-claim arg-max & central claims covered \\
Relation strength & max support, max refutation, mean neutral & extrema and means & strongest local signals \\
Uncertainty & entropy, disagreement, conflict & dispersion statistics & unstable or conflicting evidence \\
Retrieval & retrieval uncertainty (optional) & retriever side-info & retriever confidence \\
\bottomrule
\end{tabular}
\end{table}

\begin{algorithm}[t]
\caption{SURE-RAG inference}
\label{alg:sure-rag}
\small
\begin{algorithmic}[1]
\REQUIRE Question $q$, answer $a$, evidence $E$, verifier $h_\theta$, aggregator $g_\phi$
\REQUIRE Selective threshold $\tau$ and risk weight $\beta$
\STATE Represent $a$ as claims $C(a)$
\FOR{each claim $c_j$ and passage $e_i$}
    \STATE $\mathbf{p}_{ij}\leftarrow h_\theta(q,c_j,e_i)$
\ENDFOR
\STATE Build $\mathbf{z}(q,a,E)$ from Table~\ref{tab:feature-blocks}
\STATE $\boldsymbol{\pi}\leftarrow g_\phi(\mathbf{z}(q,a,E))$
\STATE $\hat{y}\leftarrow \arg\max_y \pi_y$
\STATE $u\leftarrow$ uncertainty penalty
\STATE $s\leftarrow \pi_{\textsc{Supported}}-\beta u$
\IF{$\hat{y}=\textsc{Supported}$ \textbf{and} $s\geq\tau$}
    \STATE $d\leftarrow\textsc{Answer}$
\ELSE
    \STATE $d\leftarrow\textsc{Abstain}$
\ENDIF
\STATE \textbf{return} $\hat{y}$ and $d$
\end{algorithmic}
\end{algorithm}

\subsection{Selective Answering and Calibration}

For selective answering, SURE-RAG treats $P(\textsc{Supported})$ as $P(\textsc{Safe})$ and computes
\begin{equation}
s(q,a,E)=P(\textsc{Safe})-\beta u(q,a,E),
\label{eq:selective-score}
\end{equation}
where $u$ combines predictive entropy, evidence disagreement, conflict, coverage deficit, and optional retrieval uncertainty. The system answers only when the predicted label is \textsc{Supported} and $s\geq\tau$; otherwise it abstains.

All hyperparameters (post-hoc calibration parameters \cite{platt1999probabilistic,guo2017calibration}, threshold $\tau$, and risk weight $\beta$) are fit on the development split with no test labels used at any stage. Because calibration and selective thresholding can alter the final decision, we report the calibrated variant separately from raw SURE-RAG and measure binary ECE on the safe/unsafe decision. To probe whether construction-derived retrieval metadata encodes shortcut signals, we further report no-oracle variants that drop retrieval scores or replace them with BM25-derived scores~\cite{robertson2009bm25}.
\section{Experimental Setup}
\label{sec:experimental-setup}

We evaluate SURE-RAG along four axes: controlled evidence sufficiency verification, selective answering, artifact robustness, and external transfer. The main experiment uses HotpotQA-RAG v3; SciFact is reported only as a diagnostic pilot.

\subsection{Datasets}

\subsubsection{SciFact Diagnostic Pilot}

We first adapt SciFact \cite{wadden2020scifact} into a controlled RAG-style setting by treating each claim as the answer under a generic verification question. Full evidence uses gold rationales, and partial, irrelevant, and hard-insufficient variants remove or replace evidence; variants of the same claim share a group ID and stay in the same split. This pilot exhibited severe shortcut risk: evidence-only, length-only, and overlap-only baselines could match or exceed intended model behavior, so SciFact is reported only as a cautionary control.

\subsubsection{HotpotQA-RAG v3}

Our main benchmark is built from HotpotQA \cite{yang2018hotpotqa}, whose multi-hop structure naturally supports partial-evidence conditions. HotpotQA-RAG v3 contains 4{,}026 examples grouped into 900 question groups, with splits group-disjoint by original question ID. For full, partial, hard-insufficient, and irrelevant variants only the evidence changes; for natural refutation, the answer is perturbed while the original supporting evidence is retained.

\textbf{Full evidence} contains the required supporting facts plus distractors (labeled \textsc{Supported}). \textbf{Partial evidence} removes at least one necessary supporting fact, \textbf{hard-insufficient evidence} is topically related or high-overlap but lacks the required relation or reasoning step, and \textbf{irrelevant evidence} contains only distractors; these three are labeled \textsc{Insufficient}. \textbf{Natural refutation} perturbs the answer through entity swaps, number changes, date shifts, or yes/no flips while retaining evidence that contradicts the perturbed answer (labeled \textsc{Refuted}); prefix-negation templates such as ``not [answer]'' are banned.

\paragraph{Pair-level labels.}
Pair-level labels are derived from construction metadata, not copied from answer-level labels. HotpotQA supporting facts are labeled \textsc{Support} against the original answer; in natural refutation variants, the same supporting facts become \textsc{Refute} against the perturbed answer. Distractors, irrelevant passages, and high-overlap non-support passages are labeled \textsc{Neutral}. The benchmark yields 20{,}130 claim-evidence pairs: 3{,}292 support, 2{,}202 refute, and 14{,}636 neutral, with no unknown labels.

\subsubsection{External HaluBench Transfer}

We evaluate external transfer on HaluBench \cite{patronusai2024halubench} as a binary safe/unsafe task, since its labels do not distinguish \textsc{Refuted} from \textsc{Insufficient}; no thresholds are tuned on HaluBench test labels.

\subsection{Models and Baselines}

Shortcut baselines audit dataset artifacts: majority, hypothesis-only, evidence-only, length-only, overlap-only, and concat TF-IDF. They are not semantic competitors; if they approach SURE-RAG, the result is treated as shortcut-prone.

Neural pooling baselines use the same DeBERTa-v3-base pair verifier as SURE-RAG. Max-pooling takes the maximum relation probability across passages; mean-pooling averages them; top-$k$ pooling averages the top-scoring signals.

We also train a strong answer-level DeBERTa concat cross-encoder that receives the question, answer, and concatenated evidence and directly predicts the three-way sufficiency label, testing whether SURE-RAG's aggregation remains competitive with a stronger but less interpretable baseline.

SURE-RAG variants include the full model, the calibrated model, pooling-only aggregation, and ablations that individually remove claim coverage, evidence disagreement, conflict, or retrieval uncertainty.

Finally, we evaluate GPT-4o~\cite{openai2024gpt4o} as an LLM judge. The model receives only the question, answer, and evidence, is instructed to ground its decision strictly in the provided evidence, and returns a JSON label with no chain-of-thought. Gold labels and any SURE-RAG or baseline predictions are excluded from the prompt. We use one deterministic call per example at temperature 0 and evaluate on matched sampled IDs, so SURE-RAG and GPT-4o are compared on identical inputs.

\subsection{Training, Metrics, and Diagnostics}

For each neural seed, we train the pair verifier on claim-evidence pairs and cache its predictions for all splits. SURE-RAG aggregation is trained on the training split and tuned on the development split. Main HotpotQA-RAG v3 neural results are averaged over seeds 13, 21, and 42.

For three-way classification, we report Macro-F1 and per-class F1; class-wise behavior is more diagnostic than accuracy alone for sufficiency. For binary safety, we report safe-F1 and unsafe-F1, with \textsc{Supported} mapped to safe. For selective answering, we report risk at fixed coverage; fixed-risk operating points (coverage at fixed risk) are reported only as diagnostics, since they were conservative in our setting. We also report binary expected calibration error (ECE) on the safe/unsafe decision, with confidence $\max(P(\textsc{Safe}),1-P(\textsc{Safe}))$.

Trustworthiness diagnostics include artifact ratio, counterfactual evidence swaps, GPT-4o semantic refutation audit, no-oracle retrieval-score checks, and HaluBench transfer. The artifact ratio is the best shortcut-baseline Macro-F1 divided by SURE-RAG Macro-F1. Counterfactual swaps test whether $P(\textsc{Supported})$ is higher for full evidence than for degraded evidence under the same question and answer.

\subsection{Implementation and Reproducibility}

Table~\ref{tab:impl-details} summarizes key implementation settings. Subject to source dataset licenses, we plan to release the processed JSONL files, group-disjoint split IDs, configurations, model prediction files, and evaluation scripts.

\begin{table}[t]
\centering
\small
\setlength{\tabcolsep}{3pt}
\caption{Implementation details for SURE-RAG experiments.}
\label{tab:impl-details}
\begin{tabular}{ll}
\toprule
Component & Setting \\
\midrule
Pair verifier & DeBERTa-v3-base cross-encoder \\
Aggregator & logistic regression (answer-level) \\
Aggregator input & relation, coverage, uncertainty, retrieval blocks \\
Calibration & development-set post-hoc calibration \\
Selective rule & $P(\textsc{Safe})-\beta u$, dev-selected $\tau,\beta$ \\
Seeds & 13, 21, 42 \\
Split policy & group-disjoint by question ID \\
Pair labels & metadata-derived (no unknown labels) \\
LLM judge & GPT-4o, temperature 0, JSON-only output \\
\bottomrule
\end{tabular}
\end{table}

\section{Results}
\label{sec:results}

\subsection{Diagnostic SciFact Results}

On SciFact, hypothesis-only, evidence-only, length-only, and overlap-only baselines reach artifact ratios of 0.9195, 1.2945, 1.2117, and 1.4052; multiple shortcut signals match or exceed semantic models. This confirms SciFact's cautionary status and motivates the artifact-aware protocol used throughout the rest of the paper.

\subsection{Main HotpotQA-RAG v3 Results}

Table~\ref{tab:main-results} reports HotpotQA-RAG v3 results averaged over three neural seeds. Calibrated SURE-RAG achieves 0.9075 Macro-F1, with raw SURE-RAG at 0.8951 $\pm$ 0.0069, well above DeBERTa mean-pooling (0.6516) and on par with the strong concat cross-encoder (0.8888 $\pm$ 0.0109).

\begin{table*}[t]
\centering
\small
\setlength{\tabcolsep}{5pt}
\caption{Main HotpotQA-RAG v3 classification results. Macro-F1 is mean $\pm$ standard deviation over three seeds for multi-seed neural models. Per-class F1 values are seed-averaged means.}
\label{tab:main-results}
\begin{tabular}{lccccc}
\toprule
Model & $n$ & Macro-F1 & Supp.-F1 & Ref.-F1 & Insuff.-F1 \\
\midrule
Majority & 1 & 0.1848 & 0.0000 & 0.0000 & 0.5545 \\
Concat TF-IDF & 1 & 0.3501 & 0.1558 & 0.3095 & 0.5848 \\
Overlap-only & 1 & 0.6101 & 0.6394 & 0.6775 & 0.5133 \\
\midrule
DeBERTa max-pool & 3 & 0.5948 $\pm$ 0.0639 & 0.3203 & 0.7927 & 0.6715 \\
DeBERTa mean-pool & 3 & 0.6516 $\pm$ 0.0595 & 0.5444 & 0.7307 & 0.6798 \\
DeBERTa top-$k$ pool & 3 & 0.6444 $\pm$ 0.0603 & 0.5417 & 0.7136 & 0.6778 \\
\midrule
Concat cross-encoder & 3 & 0.8888 $\pm$ 0.0109 & 0.8496 & 0.9437 & 0.8731 \\
SURE pooling-only & 3 & 0.8383 $\pm$ 0.0115 & 0.7788 & 0.9393 & 0.7967 \\
SURE-RAG & 3 & 0.8951 $\pm$ 0.0069 & 0.8465 & 0.9508 & 0.8878 \\
SURE-RAG calibrated & 3 & \textbf{0.9075 $\pm$ 0.0060} & \textbf{0.8662} & \textbf{0.9569} & \textbf{0.8994} \\
\bottomrule
\end{tabular}
\end{table*}

\subsection{Artifact and No-Oracle Checks}

Table~\ref{tab:trust-checks} summarizes trustworthiness checks. The strongest shortcut baseline is overlap-only, with artifact ratio 0.6816, below our severe-risk threshold but indicating moderate shortcut risk; HotpotQA-RAG v3 is therefore a controlled benchmark.

No-oracle checks test whether construction-derived retrieval scores drive the result. Dropping these scores yields 0.8370 Macro-F1 and replacing them with BM25-derived scores yields 0.8347, and both remain well above DeBERTa mean-pooling (0.6516). On the refutation audit, GPT-4o validates 260 of 300 sampled candidates (valid-refutation rate 0.8667); on this audited subset, SURE-RAG obtains 0.9370 refuted-F1.

\begin{table}[t]
\centering
\small
\caption{Trustworthiness checks. No-oracle results show that SURE-RAG remains stronger than pooling after removing construction-derived retrieval scores.}
\label{tab:trust-checks}
\begin{tabular}{lc}
\toprule
Check & Result \\
\midrule
Max artifact ratio & 0.6816 \\
Prefix-not rate & 0.0000 \\
GPT-4o valid-refutation rate & 0.8667 \\
SURE, drop retrieval score & 0.8370 Macro-F1 \\
SURE, BM25 retrieval score & 0.8347 Macro-F1 \\
DeBERTa mean-pool & 0.6516 Macro-F1 \\
\bottomrule
\end{tabular}
\end{table}

\subsection{Selective Answering}

Table~\ref{tab:selective-results} reports selective-answering risk. At 30\% coverage, SURE-RAG reaches Risk@30 of 0.1642, a 37\% relative reduction over DeBERTa mean-pooling (0.2588) and roughly half of max-pooling (0.3183). At higher coverage the advantage narrows, so the gain is in low-coverage ranking rather than strict high-coverage risk control.

\begin{table}[t]
\centering
\small
\caption{Selective answering on HotpotQA-RAG v3. SURE-RAG improves low-coverage risk.}
\label{tab:selective-results}
\begin{tabular}{lccc}
\toprule
Model & Risk@30 & Risk@50 & Risk@70 \\
\midrule
DeBERTa max-pool & 0.3183 & 0.4054 & 0.5344 \\
DeBERTa mean-pool & 0.2588 & 0.3783 & 0.5350 \\
SURE-RAG & \textbf{0.1642} & 0.3600 & 0.5457 \\
SURE-RAG calibrated & 0.1670 & \textbf{0.3510} & 0.5493 \\
\bottomrule
\end{tabular}
\end{table}

\subsection{Calibration and Fixed-Risk Diagnostics}

On the safe/unsafe decision (Table~\ref{tab:calibration-diagnostics}), calibration reduces binary ECE from 0.0304 (raw) to 0.0198 (calibrated). This validates the post-hoc calibration step, but does not by itself improve high-coverage risk control: in our setting, strict fixed-risk thresholds were too conservative to yield useful coverage, so we report fixed-risk operating points only as diagnostics.

\begin{table}[t]
\centering
\small
\caption{Calibration diagnostics for the safe/unsafe decision on HotpotQA-RAG v3. Lower binary ECE is better.}
\label{tab:calibration-diagnostics}
\begin{tabular}{lcc}
\toprule
Model & Binary ECE $\downarrow$ & Risk@30 $\downarrow$ \\
\midrule
SURE-RAG & 0.0304 & 0.1642 \\
SURE-RAG calibrated & \textbf{0.0198} & 0.1670 \\
\bottomrule
\end{tabular}
\end{table}

\subsection{Counterfactual Evidence Swap}

For grouped examples, we compare $P(\textsc{Supported})$ under full evidence against degraded evidence with the same question and answer. Table~\ref{tab:counterfactual} shows perfect or near-perfect counterfactual sensitivity: success is 1.0 for partial, hard-insufficient, and irrelevant comparisons, and 0.9679 for refuting comparisons.

\begin{table}[t]
\centering
\small
\caption{Counterfactual evidence swap. SURE-RAG assigns higher supported probability to full evidence than to degraded evidence conditions.}
\label{tab:counterfactual}
\begin{tabular}{lcc}
\toprule
Condition pair & Mean $\Delta P_{\mathrm{sup}}$ & Success \\
\midrule
Full vs. partial & 0.4177 & 1.0000 \\
Full vs. hard-insufficient & 0.7672 & 1.0000 \\
Full vs. irrelevant & 0.7967 & 1.0000 \\
Full vs. refuting & 0.7378 & 0.9679 \\
\bottomrule
\end{tabular}
\end{table}

\subsection{GPT-4o Judge and External Transfer}

We use SURE-RAG vs.\ GPT-4o as a paired probe across the two task settings (Table~\ref{tab:llm-external}). On HotpotQA-RAG v3 (matched 438 examples), SURE-RAG reaches 0.8951 Macro-F1 versus 0.7284 for GPT-4o, showing that a task-specific verifier can outperform a general-purpose LLM judge on controlled sufficiency. On HaluBench (binary safe/unsafe, with PASS$\to$safe and FAIL$\to$unsafe), the ranking reverses: GPT-4o reaches 0.7389 unsafe-F1, while SURE-RAG reaches 0.3343 and the best pooling baseline reaches 0.4863. The error overlap is asymmetric (291 cases correct only by GPT-4o vs.\ 113 only by SURE-RAG), consistent with domain shift, longer natural answers, label mismatch, and the absence of controlled evidence conditions on HaluBench. The reversed ranking establishes that controlled sufficiency verification and natural hallucination detection are distinct problems requiring different inductive biases.

\begin{table}[t]
\centering
\small
\caption{GPT-4o judge and external HaluBench transfer on matched samples. GPT-4o is weaker than SURE-RAG on controlled HotpotQA-RAG v3 but stronger on natural HaluBench unsafe detection.}
\label{tab:llm-external}
\begin{tabular}{lcc}
\toprule
Setting & Model & Metric \\
\midrule
HotpotQA-RAG v3 & GPT-4o & 0.7284 Macro-F1 \\
HotpotQA-RAG v3 & SURE-RAG & 0.8951 Macro-F1 \\
HaluBench & GPT-4o & 0.7389 unsafe-F1 \\
HaluBench & SURE-RAG & 0.3343 unsafe-F1 \\
HaluBench & best pooling & 0.4863 unsafe-F1 \\
\bottomrule
\end{tabular}
\end{table}

\subsection{Ablation}

Table~\ref{tab:ablation-results} reports ablations. SURE-RAG outperforms pooling-only aggregation by 5.7 Macro-F1 points, confirming the value of answer-level sufficiency aggregation. The largest single contributor is retrieval uncertainty: removing it drops Macro-F1 from 0.8951 to 0.8370, although this no-retrieval variant still far exceeds the DeBERTa pooling baselines (Table~\ref{tab:main-results}). Removing disagreement, conflict, or claim coverage individually has minimal effect on this short-answer benchmark.

\begin{table}[t]
\centering
\small
\caption{Ablation results on HotpotQA-RAG v3. Aggregation improves over pooling, while individual claim-level features are not fully validated in the short-answer setting.}
\label{tab:ablation-results}
\begin{tabular}{lc}
\toprule
Model & Macro-F1 \\
\midrule
SURE-RAG calibrated & 0.9075 \\
SURE-RAG & 0.8951 \\
w/o disagreement & 0.9060 \\
w/o conflict & 0.8957 \\
w/o claim coverage & 0.8953 \\
Pooling-only & 0.8383 \\
w/o retrieval uncertainty & 0.8370 \\
DeBERTa mean-pool & 0.6516 \\
\bottomrule
\end{tabular}
\end{table}

\section{Discussion and Limitations}

The experiments establish SURE-RAG as an effective verifier for controlled evidence sufficiency.

\subsection{Interpretation}

The main result confirms that evidence sufficiency is distinct from retrieval relevance: SURE-RAG improves over pair-level pooling because it treats the retrieved context as a set-level sufficiency object rather than the most supportive-looking passage. It also matches the strong concat cross-encoder through a structured aggregation protocol, and its four feature blocks expose interpretable structure that a concat baseline does not, which makes it easier to audit shortcuts, remove construction-derived retrieval features, and derive selective-answering scores.

The trustworthiness checks support these numbers: no-oracle variants stay well above pooling, so retrieval metadata does not drive the result, and the GPT-4o audit validates most sampled refutations. With artifact risk only moderate, the headline results are strong controlled-benchmark evidence.

Selective answering yields a clear low-coverage ranking benefit, and post-hoc calibration improves binary ECE on the safe/unsafe decision; the gain concentrates at low coverage rather than providing strict high-coverage risk guarantees.

The paired probe with GPT-4o locates the task boundary: SURE-RAG dominates on controlled HotpotQA-RAG v3 and GPT-4o dominates on natural HaluBench unsafe detection, so the two settings reward different inductive biases rather than ranking models consistently.

\subsection{Limitations}

\paragraph{Benchmark scope.}
HotpotQA-RAG v3 is a controlled, QA-derived benchmark that isolates sufficiency conditions; transfer to real RAG logs and naturally generated long-form responses remains to be tested. The maximum shortcut-baseline artifact ratio is 0.6816, so the benchmark is controlled but not shortcut-free.

\paragraph{Refutation noise.}
Entity-swap perturbations occasionally create answer-type mismatches; type-preserving perturbations, entity linking, or human verification would further reduce this noise.

\paragraph{Long-form claim coverage.}
SURE-RAG is formulated at the claim level, but our experiments validate short-answer sufficiency; long-form multi-claim coverage is future work.

\paragraph{Scope and supervision.}
SURE-RAG targets controlled sufficiency verification rather than open-world hallucination detection, which would require broader data, richer labels, and more diverse training. Its GPT-4o audits and judge baseline complement rather than replace large-scale human annotation.

\paragraph{Joint passage reasoning.}
SURE-RAG aggregates local claim-evidence signals without end-to-end passage-passage reasoning, which more complex multi-hop settings may require.

\section{Conclusion}

We studied evidence sufficiency verification as a selective-answering layer for RAG. SURE-RAG predicts whether retrieved evidence supports, refutes, or is insufficient for an answer and abstains when support is absent. On controlled HotpotQA-RAG v3, SURE-RAG clearly outperforms pair-level pooling baselines, matches a strong concat cross-encoder while remaining fully auditable, survives no-oracle retrieval-score checks, and lowers unsafe-answer risk at low coverage.

These results also map the boundary of the current approach. HotpotQA-RAG v3 is controlled rather than naturalistic, artifact risk is moderate, and our validation covers short-answer rather than long-form sufficiency. The paired probe with GPT-4o further locates the task boundary: evidence sufficiency verification is a useful reliability component for RAG, but not a complete solution to open-world hallucination detection. Future work should extend the method to real RAG logs, human-audited refutation data, long-form citation-grounded answers, and risk-control methods that preserve useful coverage.

\bibliographystyle{IEEEtran}
\bibliography{references}

@inproceedings{lewis2020rag,
  title     = {Retrieval-Augmented Generation for Knowledge-Intensive {NLP} Tasks},
  author    = {Lewis, Patrick and Perez, Ethan and Piktus, Aleksandra and Petroni, Fabio and Karpukhin, Vladimir and Goyal, Naman and K{\"u}ttler, Heinrich and Lewis, Mike and Yih, Wen-tau and Rockt{\"a}schel, Tim and Riedel, Sebastian and Kiela, Douwe},
  booktitle = {Advances in Neural Information Processing Systems},
  volume    = {33},
  pages     = {9459--9474},
  year      = {2020}
}

@inproceedings{karpukhin2020dpr,
  title     = {Dense Passage Retrieval for Open-Domain Question Answering},
  author    = {Karpukhin, Vladimir and O{\u{g}}uz, Barlas and Min, Sewon and Lewis, Patrick and Wu, Ledell and Edunov, Sergey and Chen, Danqi and Yih, Wen-tau},
  booktitle = {Proceedings of the 2020 Conference on Empirical Methods in Natural Language Processing},
  pages     = {6769--6781},
  year      = {2020},
  publisher = {Association for Computational Linguistics},
  doi       = {10.18653/v1/2020.emnlp-main.550},
  url       = {https://aclanthology.org/2020.emnlp-main.550/}
}

@inproceedings{izacard2021fid,
  title     = {Leveraging Passage Retrieval with Generative Models for Open Domain Question Answering},
  author    = {Izacard, Gautier and Grave, Edouard},
  booktitle = {Proceedings of the 16th Conference of the European Chapter of the Association for Computational Linguistics: Main Volume},
  pages     = {874--880},
  year      = {2021},
  publisher = {Association for Computational Linguistics},
  doi       = {10.18653/v1/2021.eacl-main.74},
  url       = {https://aclanthology.org/2021.eacl-main.74/}
}

@inproceedings{thorne2018fever,
  title     = {{FEVER}: A Large-scale Dataset for Fact Extraction and {VER}ification},
  author    = {Thorne, James and Vlachos, Andreas and Christodoulopoulos, Christos and Mittal, Arpit},
  booktitle = {Proceedings of the 2018 Conference of the North American Chapter of the Association for Computational Linguistics: Human Language Technologies},
  pages     = {809--819},
  year      = {2018},
  publisher = {Association for Computational Linguistics},
  doi       = {10.18653/v1/N18-1074},
  url       = {https://aclanthology.org/N18-1074/}
}

@inproceedings{wadden2020scifact,
  title     = {Fact or Fiction: Verifying Scientific Claims},
  author    = {Wadden, David and Lin, Shanchuan and Lo, Kyle and Wang, Lucy Lu and van Zuylen, Madeleine and Cohan, Arman and Hajishirzi, Hannaneh},
  booktitle = {Proceedings of the 2020 Conference on Empirical Methods in Natural Language Processing},
  pages     = {7534--7550},
  year      = {2020},
  publisher = {Association for Computational Linguistics},
  doi       = {10.18653/v1/2020.emnlp-main.609},
  url       = {https://aclanthology.org/2020.emnlp-main.609/}
}

@inproceedings{aly2021feverous,
  title     = {{FEVEROUS}: Fact Extraction and Verification over Unstructured and Structured Information},
  author    = {Aly, Rami and Guo, Zhijiang and Schlichtkrull, Michael and Thorne, James and Vlachos, Andreas and Christodoulopoulos, Christos and Cocarascu, Oana and Mittal, Arpit},
  booktitle = {Proceedings of the Neural Information Processing Systems Track on Datasets and Benchmarks},
  year      = {2021},
  url       = {https://datasets-benchmarks-proceedings.neurips.cc/paper/2021/hash/68d30a9594728bc39aa24be94b319d21-Abstract-round2.html}
}

@inproceedings{schuster2021vitaminc,
  title     = {Get Your Vitamin {C}! Robust Fact Verification with Contrastive Evidence},
  author    = {Schuster, Tal and Fisch, Adam and Barzilay, Regina},
  booktitle = {Proceedings of the 2021 Conference of the North American Chapter of the Association for Computational Linguistics: Human Language Technologies},
  pages     = {624--643},
  year      = {2021},
  publisher = {Association for Computational Linguistics},
  doi       = {10.18653/v1/2021.naacl-main.52},
  url       = {https://aclanthology.org/2021.naacl-main.52/}
}

@inproceedings{yang2018hotpotqa,
  title     = {{HotpotQA}: A Dataset for Diverse, Explainable Multi-hop Question Answering},
  author    = {Yang, Zhilin and Qi, Peng and Zhang, Saizheng and Bengio, Yoshua and Cohen, William W. and Salakhutdinov, Ruslan and Manning, Christopher D.},
  booktitle = {Proceedings of the 2018 Conference on Empirical Methods in Natural Language Processing},
  pages     = {2369--2380},
  year      = {2018},
  publisher = {Association for Computational Linguistics},
  doi       = {10.18653/v1/D18-1259},
  url       = {https://aclanthology.org/D18-1259/}
}

@inproceedings{es2024ragas,
  title     = {{RAGA}s: Automated Evaluation of Retrieval Augmented Generation},
  author    = {Es, Shahul and James, Jithin and Espinosa Anke, Luis and Schockaert, Steven},
  booktitle = {Proceedings of the 18th Conference of the European Chapter of the Association for Computational Linguistics: System Demonstrations},
  pages     = {150--158},
  year      = {2024},
  address   = {St. Julians, Malta},
  publisher = {Association for Computational Linguistics},
  url       = {https://aclanthology.org/2024.eacl-demo.16/}
}

@inproceedings{saadfalcon2024ares,
  title     = {{ARES}: An Automated Evaluation Framework for Retrieval-Augmented Generation Systems},
  author    = {Saad-Falcon, Jon and Khattab, Omar and Potts, Christopher and Zaharia, Matei},
  booktitle = {Proceedings of the 2024 Conference of the North American Chapter of the Association for Computational Linguistics: Human Language Technologies},
  pages     = {338--354},
  year      = {2024},
  address   = {Mexico City, Mexico},
  publisher = {Association for Computational Linguistics},
  doi       = {10.18653/v1/2024.naacl-long.20},
  url       = {https://aclanthology.org/2024.naacl-long.20/}
}

@inproceedings{gao2023alce,
  title     = {Enabling Large Language Models to Generate Text with Citations},
  author    = {Gao, Tianyu and Yen, Howard and Yu, Jiatong and Chen, Danqi},
  booktitle = {Proceedings of the 2023 Conference on Empirical Methods in Natural Language Processing},
  pages     = {6465--6488},
  year      = {2023},
  publisher = {Association for Computational Linguistics},
  doi       = {10.18653/v1/2023.emnlp-main.398},
  url       = {https://aclanthology.org/2023.emnlp-main.398/}
}

@inproceedings{min2023factscore,
  title     = {{FActScore}: Fine-grained Atomic Evaluation of Factual Precision in Long Form Text Generation},
  author    = {Min, Sewon and Krishna, Kalpesh and Lyu, Xinxi and Lewis, Mike and Yih, Wen-tau and Koh, Pang Wei and Iyyer, Mohit and Zettlemoyer, Luke and Hajishirzi, Hannaneh},
  booktitle = {Proceedings of the 2023 Conference on Empirical Methods in Natural Language Processing},
  pages     = {12076--12100},
  year      = {2023},
  publisher = {Association for Computational Linguistics},
  doi       = {10.18653/v1/2023.emnlp-main.741},
  url       = {https://aclanthology.org/2023.emnlp-main.741/}
}

@inproceedings{manakul2023selfcheckgpt,
  title     = {{SelfCheckGPT}: Zero-Resource Black-Box Hallucination Detection for Generative Large Language Models},
  author    = {Manakul, Potsawee and Liusie, Adian and Gales, Mark J. F.},
  booktitle = {Proceedings of the 2023 Conference on Empirical Methods in Natural Language Processing},
  pages     = {9004--9017},
  year      = {2023},
  publisher = {Association for Computational Linguistics},
  doi       = {10.18653/v1/2023.emnlp-main.557},
  url       = {https://aclanthology.org/2023.emnlp-main.557/}
}

@inproceedings{niu2024ragtruth,
  title     = {{RAGT}ruth: A Hallucination Corpus for Developing Trustworthy Retrieval-Augmented Language Models},
  author    = {Niu, Cheng and Wu, Yuanhao and Zhu, Juno and Xu, Siliang and Shum, KaShun and Zhong, Randy and Song, Juntong and Zhang, Tong},
  booktitle = {Proceedings of the 62nd Annual Meeting of the Association for Computational Linguistics},
  pages     = {10862--10878},
  year      = {2024},
  address   = {Bangkok, Thailand},
  publisher = {Association for Computational Linguistics},
  doi       = {10.18653/v1/2024.acl-long.585},
  url       = {https://aclanthology.org/2024.acl-long.585/}
}

@inproceedings{he2023debertav3,
  title     = {{DeBERTaV3}: Improving {DeBERTa} using {ELECTRA}-Style Pre-Training with Gradient-Disentangled Embedding Sharing},
  author    = {He, Pengcheng and Gao, Jianfeng and Chen, Weizhu},
  booktitle = {The Eleventh International Conference on Learning Representations (ICLR)},
  year      = {2023},
  url       = {https://openreview.net/forum?id=sE7-XhLxHA}
}

@article{robertson2009bm25,
  title   = {The Probabilistic Relevance Framework: {BM25} and Beyond},
  author  = {Robertson, Stephen and Zaragoza, Hugo},
  journal = {Foundations and Trends in Information Retrieval},
  volume  = {3},
  number  = {4},
  pages   = {333--389},
  year    = {2009},
  doi     = {10.1561/1500000019}
}

@article{chow1970reject,
  title   = {On Optimum Recognition Error and Reject Tradeoff},
  author  = {Chow, Chi-Keung},
  journal = {IEEE Transactions on Information Theory},
  volume  = {16},
  number  = {1},
  pages   = {41--46},
  year    = {1970},
  doi     = {10.1109/TIT.1970.1054406}
}

@inproceedings{geifman2017selective,
  title     = {Selective Classification for Deep Neural Networks},
  author    = {Geifman, Yonatan and El-Yaniv, Ran},
  booktitle = {Advances in Neural Information Processing Systems},
  volume    = {30},
  year      = {2017},
  url       = {https://papers.nips.cc/paper_files/paper/2017/hash/4a8423d5e91fda00bb7e46540e2b0cf1-Abstract.html}
}

@inproceedings{guo2017calibration,
  title     = {On Calibration of Modern Neural Networks},
  author    = {Guo, Chuan and Pleiss, Geoff and Sun, Yu and Weinberger, Kilian Q.},
  booktitle = {Proceedings of the 34th International Conference on Machine Learning},
  pages     = {1321--1330},
  year      = {2017},
  publisher = {PMLR},
  url       = {https://proceedings.mlr.press/v70/guo17a.html}
}

@incollection{platt1999probabilistic,
  title     = {Probabilistic Outputs for Support Vector Machines and Comparisons to Regularized Likelihood Methods},
  author    = {Platt, John C.},
  booktitle = {Advances in Large Margin Classifiers},
  editor    = {Smola, Alexander J. and Bartlett, Peter and Sch{\"o}lkopf, Bernhard and Schuurmans, Dale},
  pages     = {61--74},
  year      = {1999},
  publisher = {MIT Press}
}

@book{vovk2005algorithmic,
  title     = {Algorithmic Learning in a Random World},
  author    = {Vovk, Vladimir and Gammerman, Alexander and Shafer, Glenn},
  year      = {2005},
  publisher = {Springer}
}

@article{angelopoulos2021conformal,
  title   = {A Gentle Introduction to Conformal Prediction and Distribution-Free Uncertainty Quantification},
  author  = {Angelopoulos, Anastasios N. and Bates, Stephen},
  journal = {arXiv preprint arXiv:2107.07511},
  year    = {2021},
  url     = {https://arxiv.org/abs/2107.07511}
}

@misc{patronusai2024halubench,
  title        = {{HaluBench}},
  author       = {{Patronus AI}},
  year         = {2024},
  howpublished = {\url{https://huggingface.co/datasets/PatronusAI/HaluBench}},
  note         = {Hugging Face dataset}
}

@inproceedings{jiang2024rora,
  title = {{RORA}: Robust Free-Text Rationale Evaluation},
  author = {Jiang, Zhengping and Lu, Yining and Chen, Hanjie and Khashabi, Daniel and Van Durme, Benjamin and Liu, Anqi},
  booktitle = {Proceedings of the 62nd Annual Meeting of the Association for Computational Linguistics (Volume 1: Long Papers)},
  pages = {1070--1087},
  year = {2024},
  address = {Bangkok, Thailand},
  publisher = {Association for Computational Linguistics},
  doi = {10.18653/v1/2024.acl-long.60},
  url = {https://aclanthology.org/2024.acl-long.60/}
}

@misc{openai2024gpt4o,
  title        = {{GPT-4o} System Card},
  author       = {{OpenAI}},
  year         = {2024},
  howpublished = {\url{https://openai.com/index/gpt-4o-system-card/}},
  note         = {Accessed: 2026-05-01}
}

\end{document}